\documentclass[nofootinbib,superscriptaddress,twocolumn]{revtex4-1}

\usepackage{wrapfig}
\usepackage{graphicx}
\usepackage{float}
\usepackage{subfigure}

\usepackage{color}

\usepackage{amsmath}

\usepackage[dvipsnames]{xcolor}
\usepackage[colorlinks=true,linkcolor=blue,citecolor=blue]{hyperref}

\makeatother

\begin{document}
\title{
Benchmarking projective simulation in navigation problems
}

\author{Alexey A. Melnikov}
\thanks{Corresponding author, e-mail: alexey.melnikov@uibk.ac.at}
\affiliation{Institute for Theoretical Physics, University of Innsbruck, Technikerstra{\ss }e 21a, 6020 Innsbruck, Austria}

\author{Adi Makmal}
\affiliation{Institute for Theoretical Physics, University of Innsbruck, Technikerstra{\ss }e 21a, 6020 Innsbruck, Austria}

\author{Hans J. Briegel}
\affiliation{Institute for Theoretical Physics, University of Innsbruck, Technikerstra{\ss }e 21a, 6020 Innsbruck, Austria}
\affiliation{Department of Philosophy, University of Konstanz, Fach 17, 78457 Konstanz, Germany}

\begin{abstract}
Projective simulation (PS) is a model for intelligent agents with a deliberation capacity that is based on episodic memory. The model has been shown to provide a flexible framework for constructing reinforcement-learning agents, and it allows for quantum mechanical generalization, which leads to a speed-up in deliberation time. PS agents have been applied successfully in the context of complex skill learning in robotics, and in the design of state-of-the-art quantum experiments. In this paper, we study the performance of projective simulation in two benchmarking problems in navigation, namely the grid world and the mountain car problem. The performance of PS is compared to standard tabular reinforcement learning approaches, Q-learning and SARSA. Our comparison demonstrates that the performance of PS and standard learning approaches are qualitatively and quantitatively similar, while it is much easier to choose optimal model parameters in case of projective simulation, with a reduced computational effort of one to two orders of magnitude. Our results show that the projective simulation model stands out for its simplicity in terms of the number of model parameters, which makes it simple to set up the learning agent in unknown task environments.
\end{abstract}

\maketitle

\section{Introduction}

Projective simulation (PS) is a physical approach to agency and artificial intelligence~\cite{russell2010artificial}, which was first introduced in ref.~\cite{briegel2012projective}. The PS model has been successfully applied to problems in reinforcement learning (RL)~\cite{sutton1998reinforcement,wiering2012reinforcement}, ranging from textbook problems~\cite{briegel2012projective,mautner2013projective} to applications of complex skill learning in advanced robotics~\cite{simon2016,simon2017skill}. The model was also successfully used in a domain of quantum physics problems: for adaptive quantum computation~\cite{tiersch2015adaptive} and for designing complex quantum experiments~\cite{melnikov2017active}.

The PS model is considered to be promising for several reasons. First, the internal deliberation dynamics is entirely based on a process of random walks on graphs~\cite{briegel2012projective}. This process is well-studied in probability theory and physics, as well as in the context of randomized algorithms~\cite{randomized_algorithms_1995}, which makes it easier to analytically analyze the convergence properties of the model in certain environments~\cite{melnikov2017projective}. The stochastic internal dynamics of the PS model also highlights the possibility of physical, rather than computational, realizations of learning agents. Second, the PS approach provides a clear route for the construction of quantum-enhanced reinforcement learning agents~\cite{dunjko2016quantum,dunjko2017advances}. This route is based on mapping the random walk deliberation dynamics within the PS agent to a quantum walk dynamics~\cite{aharonov1993,aharonov2001quantum,venegas-andraca2012quantum}. The quantum dynamics allows quantum parallel processing in the sense that the excitation in the PS agent's memory walks in superposition. As it was shown in ref.~\cite{paparo2014quantum}, the quantum walk dynamics of the PS agent leads to a quadratic speed-up in deliberation time when compared to a behaviourally equivalent classical PS agent. The possibility of constructing these enhanced agents in physical systems was theoretically shown for quantum systems of trapped ions~\cite{dunjko2015quantum} and superconducting circuits~\cite{friis2015coherent}. The described quantum enhancement was recently demonstrated experimentally using a small-scale quantum information processor based on trapped ions~\cite{sriarunothai2017speeding}.

In parallel with the advancement of the quantum-mechanical PS model, it is important to study the relation of the model to other RL models, and to benchmark its performance. This study will also allow us to identify practically relevant applications, such as complex skill learning in robotics, which could be dramatically enhanced by quantum processing. In this paper we consider the reinforcement learning PS model and compare it to the standard RL models of Q-learning~\cite{rummery1994line} and SARSA~\cite{sutton1990integrated}. For comparison, we have chosen two standard navigation problems that are commonly used for benchmarking: the grid world~\cite{sutton1990integrated} and the mountain car problem~\cite{moore1990efficient}. A preliminary study of these navigation problems using PS was reported in~\cite{melnikov2014projective}.

The paper is organized as follows. In Section~\ref{sec:PS_model} we give a summary of the PS model and describe its basic features. Next, in Sections~\ref{sec:GridWorld} and~\ref{sec:MountainCar} we analyze the performance of the PS agent in the grid world and the mountain car problem in detail. In addition, we compare its performance to the performance of the standard basic RL algorithms -- Q-learning and SARSA. Then, in Section~\ref{sec:discussion} we summarize the results of the paper.

\section{The PS model} \label{sec:PS_model}

\begin{figure}[!ht]
\centering
\includegraphics[width=1\linewidth]{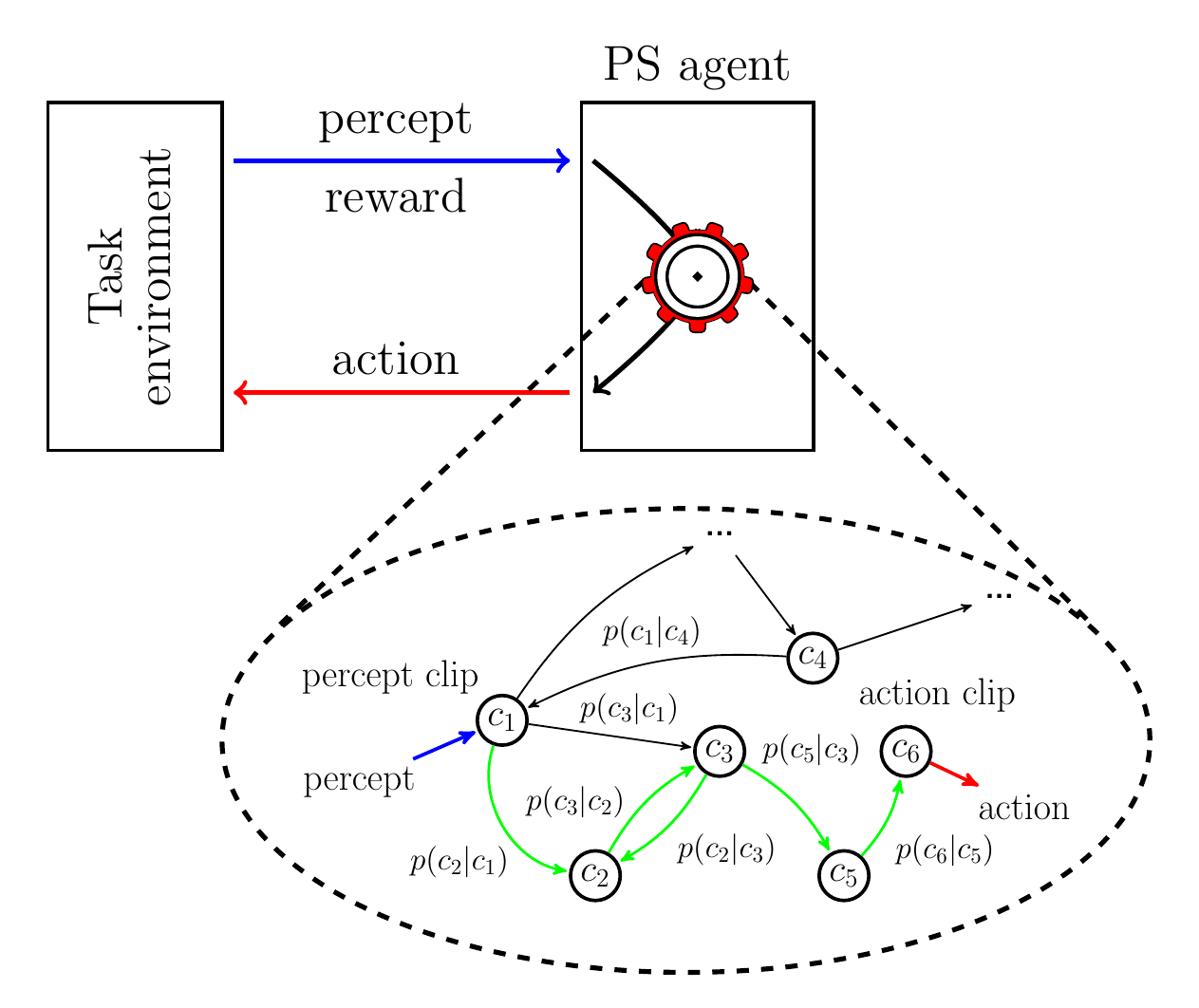}
\caption{Scheme of the PS agent that interacts with a task environment. The PS agent processes perceptual information in the network of clips. Clips are shown as circles and are labelled as $c_i$. The conditional probability to go from clip $c_i$ to clip $c_j$ is denoted as $p(c_j|c_i)$. Nonzero probabilities are shown as arrows.}
\label{fig:PSNet}
\end{figure}

PS is a framework for learning agents that interact with task environments. The interaction with a task environment is visualized schematically in Fig.~\ref{fig:PSNet}: the agent receives a percept from the environment as input (blue) and, after processing this input, outputs an action. Then, the environment evaluates the action by giving a certain reward as a part of the next input. PS differs from other learning models, such as, e.g. Q-learning and SARSA, in the way inputs (percepts, rewards) are processed. The PS agent processes percepts in a network of clips (shown at the bottom of Fig.~\ref{fig:PSNet}), which are units of the episodic memory. Clips represent remembered percepts, actions or some sequence thereof. Clips are connected via directed edges that represent possible transitions. The probability of a transition from clip $c_i$ to clip $c_j$ is a function of the edge-specific weight $h(c_i, c_j)$:
\begin{equation}
	p(c_j|c_i) = \frac{h(c_i,c_j)}{\sum_{k=1}^K h(c_i,c_k)},
\label{eq:probab1}
\end{equation}
where $K$ is the number of neighbours of the clip $c_i$. In addition, in this paper we also use the softmax function (also known as Boltzmann distribution) 
\begin{equation}
	p(c_j|c_i) = \frac{\mathrm{e}^{\beta h(c_i,c_j)}}{\sum_{k=1}^K \mathrm{e}^{\beta h(c_i,c_k)}}
\label{eq:probab2}
\end{equation}
with $\beta$ corresponding to the inverse temperature; we set $\beta=1$ throughout the paper.

The described network of clips is used for deliberation in the following way. The moment the PS agent receives a percept, the clip that corresponds to this percept is excited (blue arrow in Fig.~\ref{fig:PSNet}). Subsequently, the excitation hops in the network of clips probabilistically until it hits an action clip. An exemplary path of excitation propagation is highlighted in green in Fig.~\ref{fig:PSNet}. The excitation that hits an action clip finally triggers a real action on the environment (red arrow in Fig.~\ref{fig:PSNet}).


The learning of the PS agent is implemented by dynamic modifications of the network of clips, both by changing the topology of the PS network and by adjusting the weights of the edges. For the problems that we consider in this paper, the dynamic modifications include creation of new percept clips and adjustments of the $h$-values. The $h$-value are adjusted after each round of interaction with the environment according to the following learning rule:
\begin{eqnarray}
	h^{(t+1)}(c_i, c_j) = h^{(t)}(c_i, c_j) &-& \gamma\left(h^{(t)}(c_i, c_j)-1\right)\nonumber\\
	 &+& g^{(t+1)}(c_i, c_j)\lambda^{(t)},
\label{eq:learningRule}
\end{eqnarray}
where $0\leq\gamma\leq 1$ is a damping parameter that is responsible for forgetting, which is an important feature in changing environments~\cite{briegel2012projective,makmal2016meta,ried2017modelling}. $\lambda^{(t)}$ is a reward value that is issued by the environment at time step $t$. The so-called edge glow values $g$, are responsible for the ability to internally reward a sequence of actions instead of only the last single action. The $g$-values are updated at every step of a random walk for every edge $(c_i, c_j)$
\begin{equation}
g^{(t+1)}(c_i, c_j) = \begin{cases} 1, \mbox{ if } (c_i, c_j)\mbox{ was traversed} & \\
\left(1-\eta\right)g^{(t)}(c_i, c_j), \mbox{ otherwise}, &
\end{cases}
\label{eq:glowRule}
\end{equation}
with $0\leq\eta\leq 1$ being the edge glow damping parameter. The edge glow is essential in environments with delayed rewards (such as the grid world and the mountain car problem), in which case the $\eta$ parameter should be set to a value smaller than $1$. $g$-values determine the extent to which the previously traversed PS network edges are strengthened by the reward. As discussed in Ref.~\cite{melnikov2017projective}, the edge glow mechanism is operationally similar to the discount factor of value-based tabular RL models.

The two parameters of the PS model, $\gamma$ and $\eta$ are usually set to constant values, and it is also the case in this work. Nonetheless, the agent can benefit from having time-dependent model parameters, which can be learned autonomously by the meta-learning PS agent, which is easy to implement in the PS framework~\cite{makmal2016meta}, however at the expense of an increased learning time.

\section{The grid world problem} \label{sec:GridWorld}

\begin{figure}[!ht]
\centering
\includegraphics[width=0.9\linewidth]{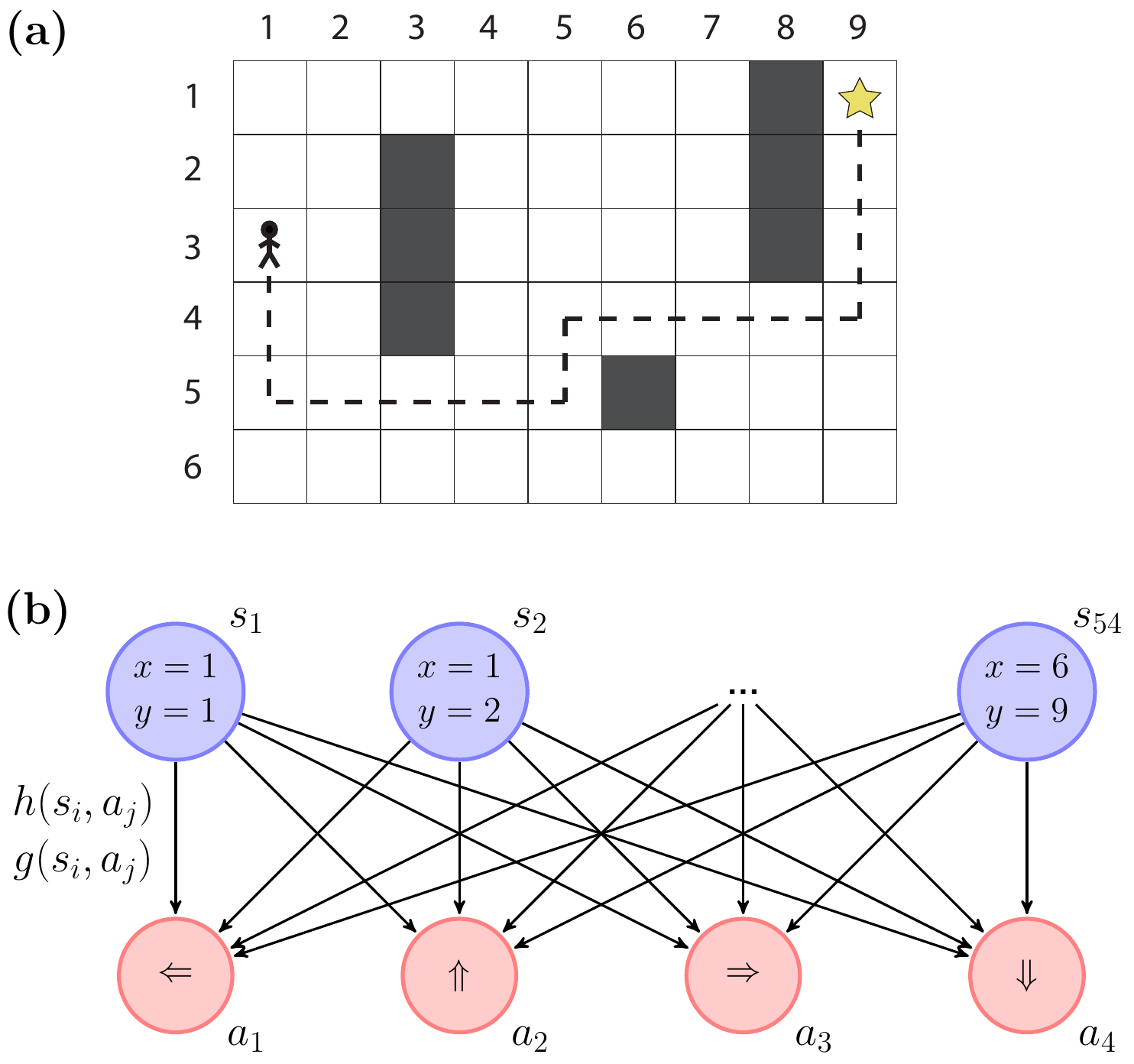}
\caption{(a) The grid world problem~\cite{sutton1990integrated}. The agent always starts a trial in the position $(3, 1)$ and then walks in the maze. The agent finishes the trial when its position is $(1, 9)$ and gets a reward (indicated by the star). An example of the shortest path to the goal is shown as a dashed line and has a length of $14$ steps. (b) The two-layered clip network that is constructed in the grid world problem. The network consists of up to $54$ percept clips (blue circles), $4$ action clips (red circles) and up to $216$ weighted edges.}
\label{fig:GridWorld}
\end{figure}

The grid world problem is a problem of navigation through a maze, which was originally described in ref.~\cite{sutton1990integrated} and later modified in many other works, see, e.g., refs~\cite{sutton1998reinforcement,mirowski2016learning,hierarchical2016maze}. In this paper, we consider the original maze of size $6\times 9$ shown in Fig.~\ref{fig:GridWorld}(a). Although the original maze is of a small size, with the help of it, several important features of an RL algorithm can be evaluated: a possibility to deal with delayed rewards and a convergence to the maximum reward per time step. The grid world environment is specified by the following rules. The agent starts in the position $(3, 1)$ of the grid at the beginning of each trial. It can make a decision to go one step up, down, left, or right. This decision brings the agent to a new location, the coordinates of which are perceived at the next time step. If the agent hits the border of the grid, or one of the walls (shaded locations in Fig.~\ref{fig:GridWorld}(a)), the position of the agent will not be changed, however a time step will be counted. The described navigation ends the moment the agent arrives at the $(1, 9)$ position of the grid (marked as the star). After the agent moves to the $(1, 9)$ location it receives a reward of $+1$, the trial ends, and the agent starts the next trial from the initial $(3, 1)$ position. The performance of the agent in this problem is measured by the number of steps that are made in each trial. A good learning agent is on average expected to improve after each trial and eventually use one of the shortest paths to reach the location with the reward.

We have solved the described grid world problem using the PS agent. The design of the underlying clip network is shown in Fig.~\ref{fig:GridWorld}(b). It consists of two types of clips: percept clips (blue) and action clips (red). At the very beginning, before the first trial, only action clips (left, up, right, down) exist. After the PS agent perceives its new $(x, y)$ coordinates -- a new percept clip is created and is connected to all actions by directed edges with initial weights $h(s_i, a_j) = 1$ and zero glow values $g(s_i, a_j) = 0$. Effectively, this procedure creates a two-layered PS network with a layer of percepts and a layer of actions. Due to the simple two-layered architecture of the clip network in the grid world problem, Eqs.~(\ref{eq:learningRule})-(\ref{eq:glowRule}) simplify to the form:
\begin{equation}
\begin{split}
g^{(t+1)} & = \left(1-\eta\right)g^{(t)}, \\
g^{(t+1)}\left(s^{(t)}, a^{(t)}\right) & = 1, \\
h^{(t+1)} & = h^{(t)} + g^{(t+1)}\lambda^{(t)}.
\end{split}
\label{eq:learningRules}
\end{equation}

\begin{figure}[!ht]
	\centering
	\includegraphics[width=1\linewidth]{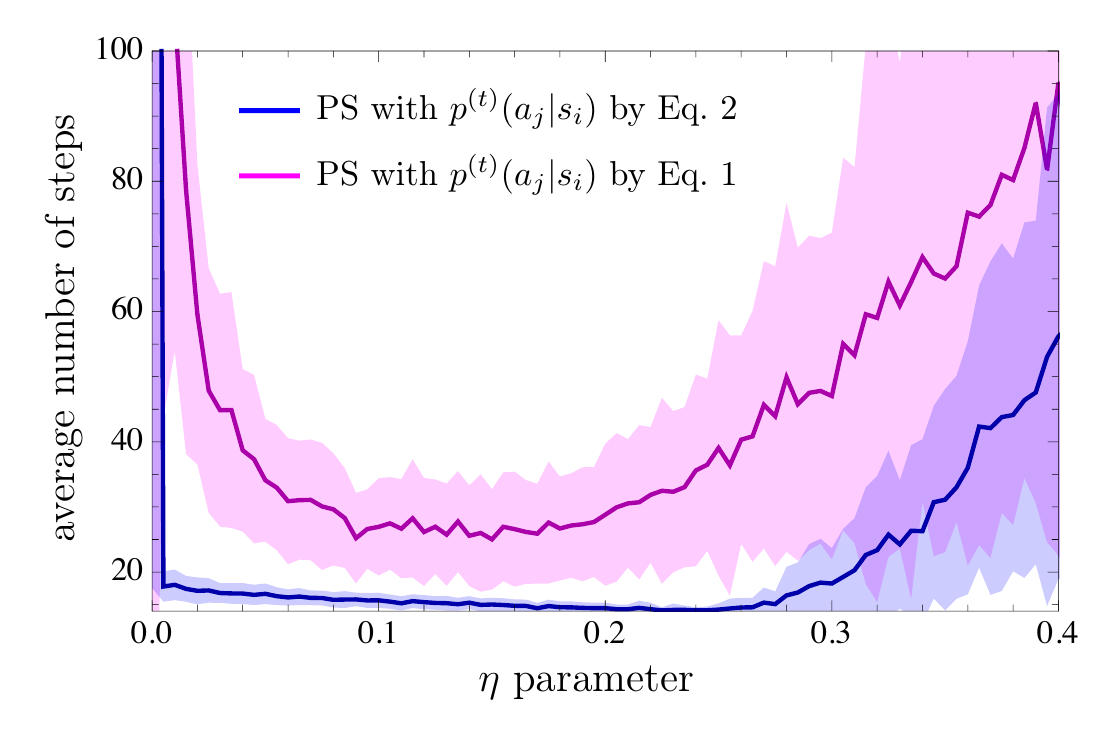}
	\caption{Performance of the PS agent in the grid world problem as a function of the $\eta$ parameter for two types of probability functions. The average number of steps is obtained after $500$ trials by averaging over $100$ agents. The shaded areas show mean squared deviation. $81$ different values for the $\eta$ parameter are uniformly distributed in the interval of $0\leq \eta\leq 0.4$ for each curve.}
	\label{fig:PSperformanceGW}
\end{figure}

We simulated the described basic PS agent in the grid world environment during $500$ trials. Each trial starts by placing the agent at the initial position and ends by the agent reaching the goal position, or by reaching the maximum number of steps in the trial, set to $10^6$ (such a large number ensures that the agent can reach the reward for all practical purposes, even by randomly walking in the maze). The results of the simulations are shown in Fig.~\ref{fig:PSperformanceGW}, where the average number of steps the agent needed in the $500$-th trial to find the rewarded site is plotted as a function of the $\eta$ parameter. We observe that PS with the softmax probability function (blue curve) from Eq.~(\ref{eq:probab2}) shows a much better performance than with the probability function from Eq.~(\ref{eq:probab1}) (magenta curve), for the entire range of considered $\eta$ parameters. This observation is explained by the fact that the softmax function enhances a small reward of $\lambda=1$ that introduces a relatively small change in the $h$-values. Hence two edges $(s_i, a_m)$ and $(s_i, a_k)$ with small difference between the weights will have a larger difference in probabilities $p(a_m|s_i)$ and $p(a_k|s_i)$ in the case of Eq.~(\ref{eq:probab2}), compared to Eq.~(\ref{eq:probab1}).

\begin{figure}[!ht]
	\centering
	\includegraphics[width=1\linewidth]{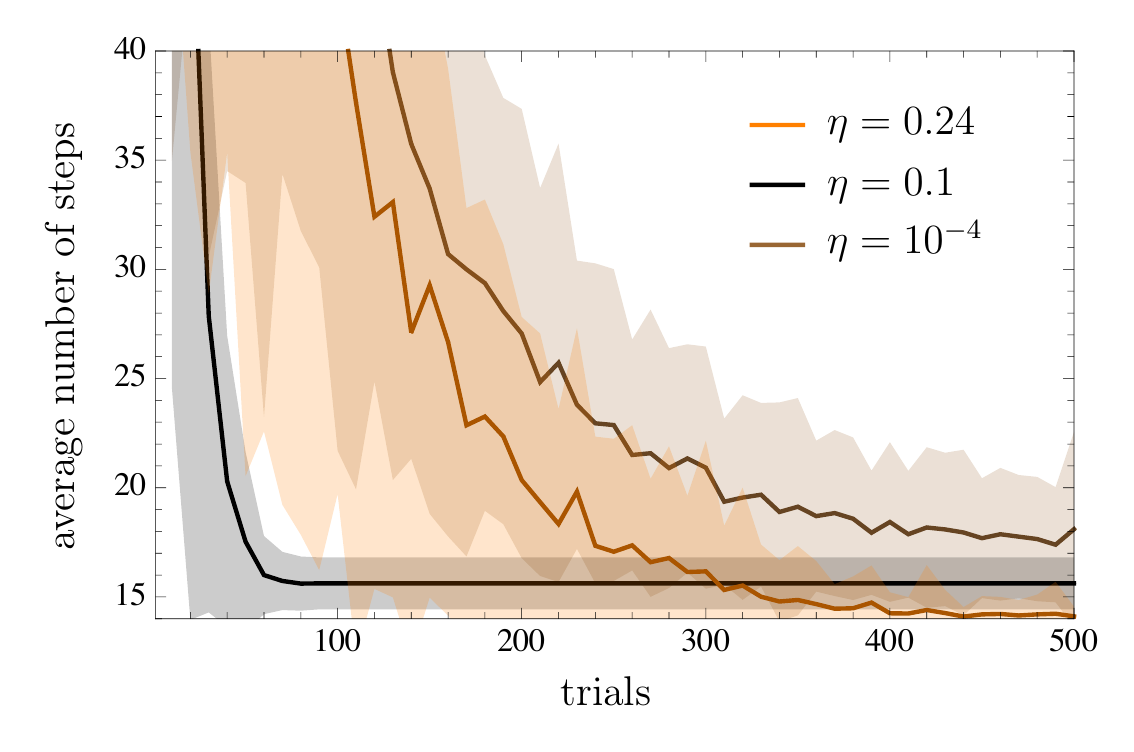}
	\caption{Learning curves of the PS agent for three different values for the $\eta$ parameter. The curves represent an average over $100$ agents of the same type. The shaded areas show mean squared deviation.}
	\label{fig:EtaStudy}
\end{figure}

The performance of PS agents with different values for the $\eta$ parameter in Fig.~\ref{fig:PSperformanceGW} indicates that there is an optimal $\eta\in [0.20, 0.25]$ parameter range for the grid world problem of the defined size. If the size of the grid world were larger, then the optimal $\eta$ parameter would have a lower value, given that the maximum number of trials is the same. The lower $\eta'$ value would lead to a larger $(1-\eta')^{l'}$ term for the path of length $l'$, which should be quantitatively comparable to the $(1-\eta)^{l}$ term in the smaller grid world with the optimal path of length $l$\footnote{The $(1-\eta)^{l}$ term enters as a coefficient in front of the reward value $\lambda$ in Eq.~(\ref{fig:PSperformanceGW}) and determines the change in the $h$-value of the edge that was traversed $l$ steps before the end of the trial.}. In other words, by changing the glow parameter one can adjust the performance of the PS agent to grid world environments of different sizes. Moreover, the glow mechanism has additional flexibility to adjust the learning time by changing the $\eta$ parameter, which is demonstrated in Fig.~\ref{fig:EtaStudy}. With $\eta = 0.1$ the agent learns relatively fast: within $50$ trials it converges to $15.6$ steps per trial. However, a better final performance can be achieved by increasing $\eta$ given that more than $50$ trials are available to the agent. For example, $\eta = 0.24$ in Fig.~\ref{fig:EtaStudy} shows a better average final performance of $14.1$ steps. If one would allow more than $500$ trials to the PS agent, the agent would show an even better performance with $\eta > 0.24$. In the opposite case, when less than $50$ trials are available to the agent, then decreasing glow parameter to a value of $\eta<0.1$ will yield an even better performance. However, the smallest possible nonzero $\eta$ is not going to give the steepest learning curve: as one can see in Fig.~\ref{fig:EtaStudy}, a small $\eta = 10^{-4}$ is not giving the fastest convergence. These simulations suggest that there is a trade-off between learning time and achieved performance after this time. Notably, depending on developer priorities, learning time or asymptotic performance can be optimized with the help of a single $\eta$ parameter.

Next we compare the performance of the PS agent with those of the Q-learning and SARSA agents. Q-learning~\cite{rummery1994line} is a standard tabular off-policy RL algorithm that learns by estimating action values $Q(s, a)$ for all states $s$ and actions $a$. The $Q$-values are updated by the following learning rule:
\begin{equation}
\begin{split}
& Q^{(t+1)}\left(s^{(t)}, a^{(t)}\right) = Q^{(t)}\left(s^{(t)}, a^{(t)}\right) + \\
& \alpha\left(\lambda^{(t)} + \mu \max_a Q^{(t)}\left(s^{(t+1)}, a\right) - Q^{(t)}\left(s^{(t)}, a^{(t)}\right)\right).
\end{split}
\label{eq:QlearningRule}
\end{equation}
SARSA~\cite{sutton1990integrated} is an on-policy tabular RL algorithm that, similar to Q-learning, estimates $Q$-values, but is defined by the following learning rule:
\begin{equation}
\begin{split}
& Q^{(t+1)}\left(s^{(t)}, a^{(t)}\right) = Q^{(t)}\left(s^{(t)}, a^{(t)}\right) + \\
& \alpha\left(\lambda^{(t)} + \mu Q^{(t)}\left(s^{(t+1)}, a^{(t+1)}\right) - Q^{(t)}\left(s^{(t)}, a^{(t)}\right)\right).
\end{split}
\label{eq:SARSARule}
\end{equation}
Both Q-learning and SARSA have the same set of parameters in their learning rules: $0 \leq \alpha \leq 1$ is the learning rate, $0 \leq \mu \leq 1$ is the discount factor and $Q^{(0)}$ is the initial action value for all state-action $\left(s, a\right)$ pairs. For these learning algorithms we consider the $\epsilon$-greedy policy: with probability $\epsilon$ the action $a^{(t+1)}$ is random, otherwise the action is chosen greedily according to the rule $a^{(t+1)} = \arg\max_a Q^{(t)}\left(s^{(t+1)}, a\right)$.

\begin{figure*}[!ht]
	\centering
	\includegraphics[width=1\linewidth]{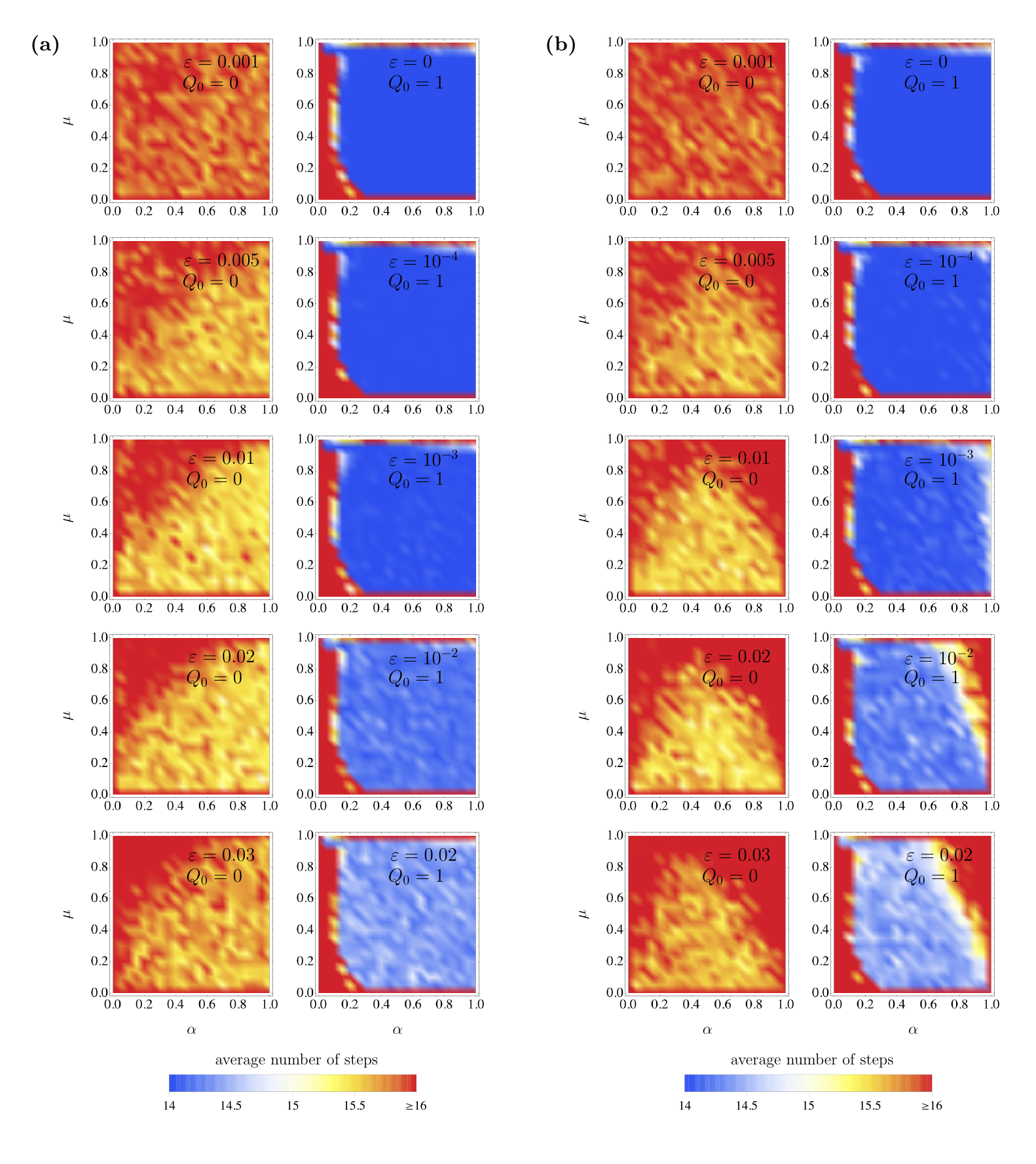}
	\caption{(a) Q-learning parameter optimization in the grid world problem. (b) SARSA parameter optimization in the grid world problem. (a)-(b) The average number of steps is calculated after $500$ trials by averaging over $100$ agents. $21$ different values for the $\mu$ and $\alpha$ parameters are uniformly distributed in the $[0, 1]$ interval for each $Q_0$ and $\epsilon$. This led to $4410$ different sets of parameter values.}
	\label{fig:QSARSA-GW}
\end{figure*}

Figure~\ref{fig:QSARSA-GW}(a) shows the results of simulations of the Q-learning agent in the grid world environment. The average number of steps needed to find the rewarded site was determined after $500$ trials for different values for the $\alpha, \mu, Q^{(0)}$ and $\epsilon$ parameters. We first observe that the initial action values of $Q^{(0)} = 1$ (the right column of Figure~\ref{fig:QSARSA-GW}(a)) are much better compared to the values of $Q^{(0)} = 0$. The reason lies in the correlation of $Q^{(0)}$ with the reward function: most of the time the agent gets no reward, which makes $Q$-values of explored actions go down to zero, and separates explored actions from unexplored ones. Such a choice of $Q^{(0)}$ effectively reduces the time needed to find the goal state for the first time. We next see that the greedy policy with $\epsilon = 0$ demonstrates the best performance, which also makes the Q-learning agent mostly independent of the learning rate $\alpha$. However, for such a performance the discount factor should be set to a value $\mu > 0.3$.

The performance of the second standard tabular RL agent that we consider in our paper, the SARSA agent, is shown in Fig.~\ref{fig:QSARSA-GW}(b). The dependence of the performance on different values of model parameters is similar to the Q-learning case, but is slightly worse for a non-optimal set of parameters. The best performance of the SARSA agent is achieved for $Q^{(0)} = 1$ and $\epsilon = 0$ like in the case of Q-learning. This, in fact, makes the two models equivalent for that choice of parameters because Eq.~(\ref{eq:QlearningRule}) and Eq.~(\ref{eq:SARSARule}) are equivalent given that the policy is greedy.

\begin{figure}[!ht]
	\centering
	\includegraphics[width=1\linewidth]{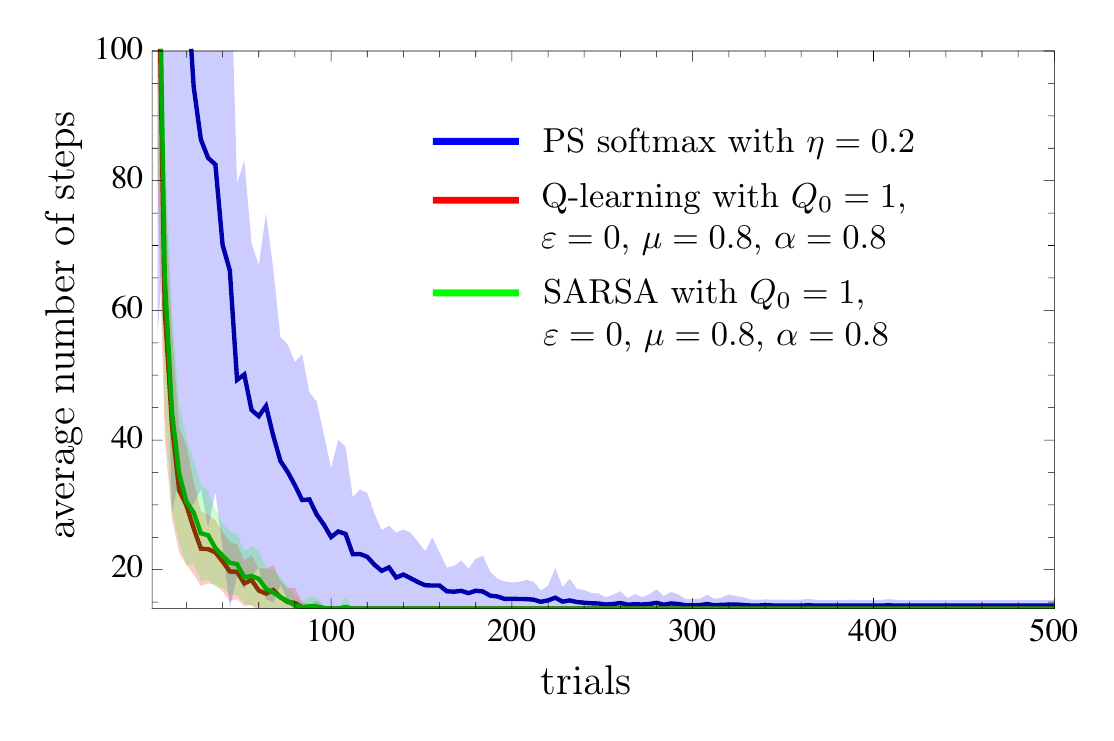}
	\caption{The learning curves of the PS (blue), Q-learning (red) and SARSA (green) agents in the grid world problem. The parameters of the agents are optimized for this problem. The average performance is calculated by averaging over $100$ agents that have the same set of parameters. The shaded areas show mean squared deviation.}
	\label{fig:GW-LearningCurves}
\end{figure}

The best parameter values that were found by simulations in Fig.~\ref{fig:PSperformanceGW} and Fig.~\ref{fig:QSARSA-GW} for PS, Q-learning and SARSA, respectively, lead to the learning curves plotted in Fig.~\ref{fig:GW-LearningCurves}. These learning curves show that all three types of RL agents improve their performance gradually and converge to the optimal behavior with $14$ steps per trial. To be more specific, PS finishes the $500$-th trial with the

\clearpage

\noindent average path length of $14.46$, Q-learning and SARSA do it in $14.00$ steps. One can see that Q-learning (red) and SARSA (green) converge to the shortest path faster than PS (blue). This however was possible only after an extensive parameter search procedure that included checking the performance of $4410$ different agents, both for Q-learning (see Fig.~\ref{fig:QSARSA-GW}(a)) and SARSA (see Fig.~\ref{fig:QSARSA-GW}(b)), compared to just $81$ different agents for PS (see Fig.~\ref{fig:PSperformanceGW}).


\section{The mountain car problem} \label{sec:MountainCar}

\begin{figure}[!ht]
	\centering
	\includegraphics[width=0.8\linewidth]{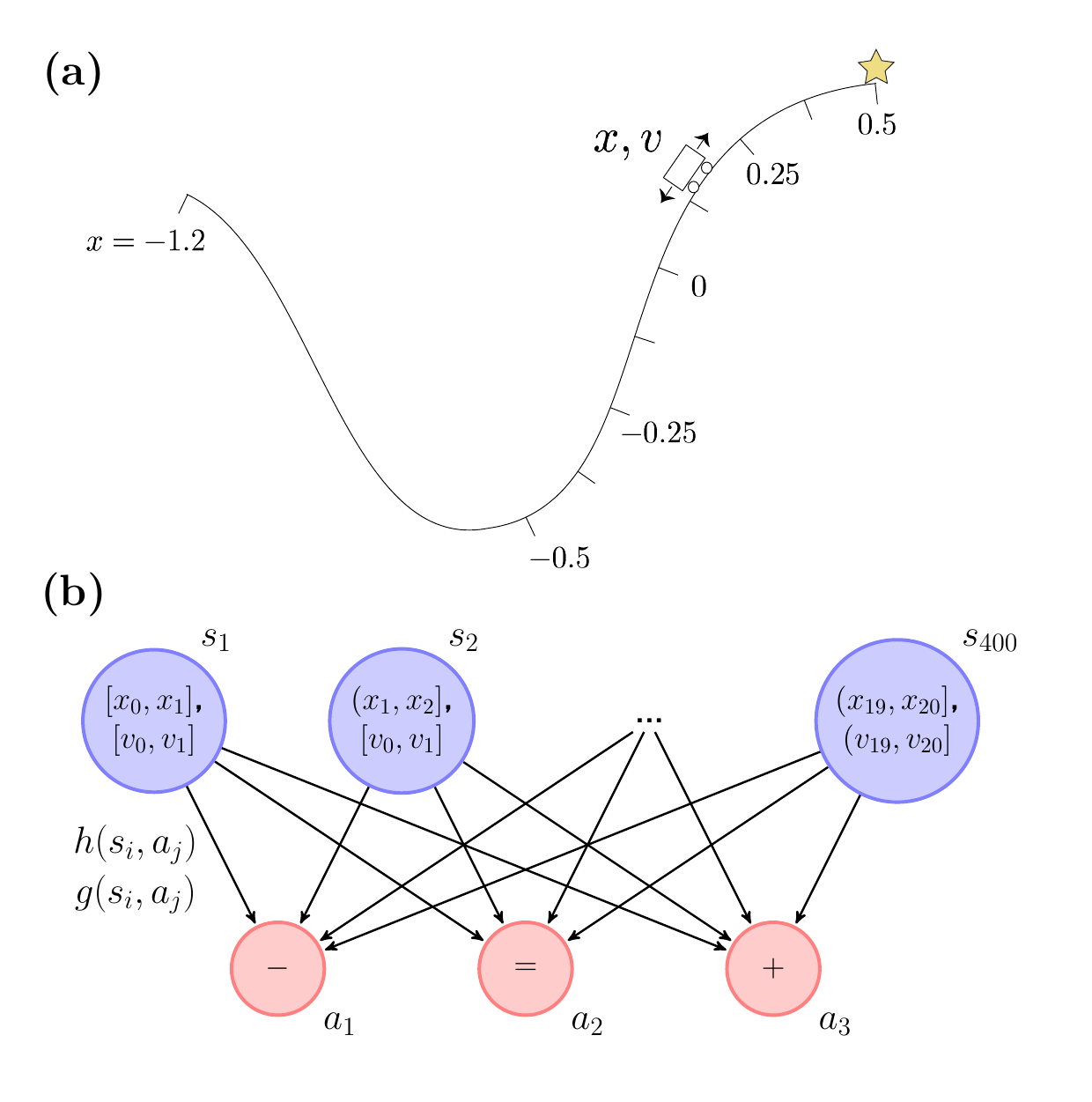}
	\caption{(a) The mountain car problem. The agent always starts in the valley and then drives the car. The agent finishes the trial when its position is $x\geq 0.5$ and gets a reward (the star). (b) The two-layered PS network used in the mountain car problem. The network consists of up to $400$ percept clips (blue circles), $3$ action clips (red circles) and up to $1200$ weighted edges.}
	\label{fig:MountainCarProblem}
\end{figure}

The second navigation problem that we consider is the mountain car problem, which was first introduced in Ref.~\cite{moore1990efficient} and is a part of a modern benchmarking framework~\cite{OpenAIpaper}. In this problem an agent is tasked to drive a car shown in Fig.~\ref{fig:MountainCarProblem}(a) and reach the goal at position $x=0.5$. The agent starts with the zero velocity at the position $x=-0.5$ close to the bottom in between two hills. The agent perceives the coordinate and the velocity of the car at each time step, $s^{(t)} = (x^{(t)}, v^{(t)})$, and is able to control the car by choosing between three available actions: accelerate to the left, accelerate to the right, or do not accelerate. An action changes the position and the velocity of the car according to the following rule~\cite{sutton1998reinforcement}:
\begin{equation}
  \begin{array}{ccl}
    v^{(t+1)} & = & v^{(t)} + 0.001 a^{(t)} - 0.0025\cos (3x^{(t)}),\\
    x^{(t+1)} & = & x^{(t)} + v^{(t+1)},
  \end{array}
\label{eq:MCstateUpdate}
\end{equation}
where $a^{(t)}\in\{-1,0,1\}$ is the action chosen at time step $t$. The velocity of the car is limited such that $-0.07\leq v\leq 0.07$. In case the velocity exceeds the absolute value of $0.07$, this value is set to $0.07$~\cite{sutton1998reinforcement}. The position is also bounded so that $-1.2\leq x \leq 0.6$. If the car tries to pass the left end of the position space, the position is set to $x=-1.2$ and velocity is reset to zero, as if the car is stopped by a wall.

\begin{figure}[!ht]
	\centering
	\includegraphics[width=1\linewidth]{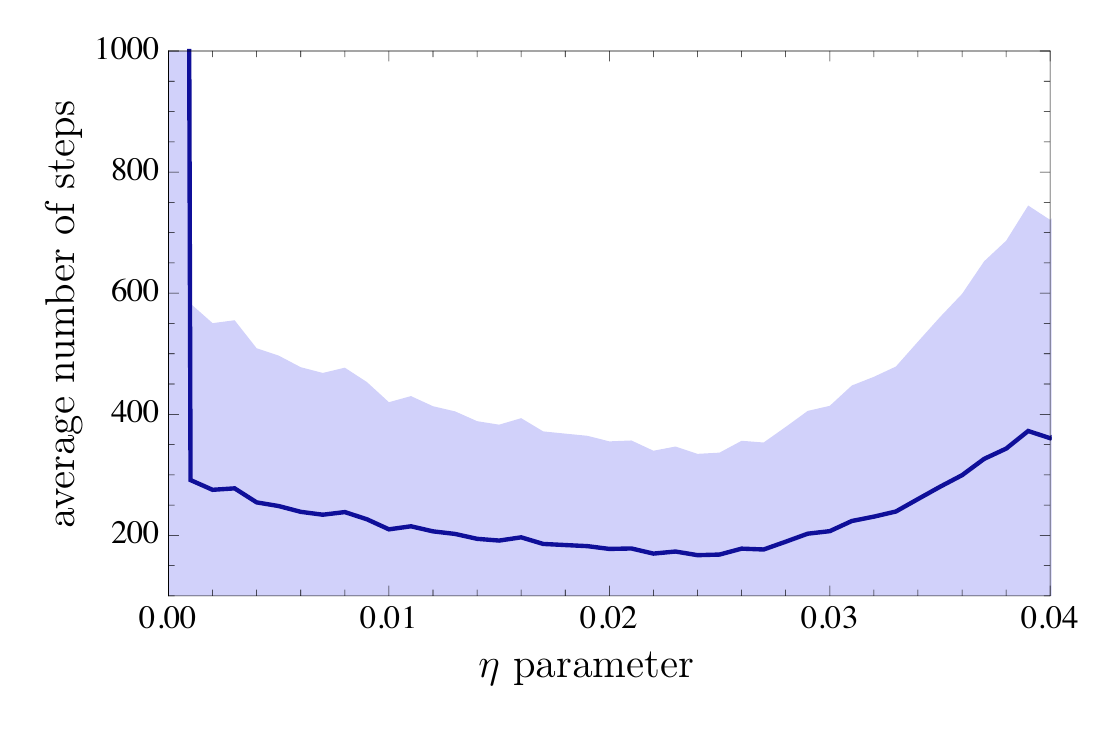}
	\caption{The average number of steps spent before reaching the goal in the mountain car problem as a function of the glow parameter of the PS agent. The average number of steps is calculated after $1000$ trials by averaging over $100$ agents. The shaded area shows mean squared deviation. $41$ different values for the $\eta$ parameter are uniformly distributed in the interval of $0\leq \eta\leq 0.04$ for each curve.}
	\label{fig:PSetaMC}
\end{figure}

The clip network that is built up by PS in the mountain car problem is shown in Fig.~\ref{fig:MountainCarProblem}(b). The network is similar to the one from the grid world problem and consists of layers of percepts (blue) and actions (red). In order to keep the percept space finite, we discretize the possible values of $(x, v)$ pairs by dividing both the position and the momentum range into $20$ equal intervals. This division limits the maximum number of clips in the PS network to $400$ percept clips and $3$ action clips. The performance of the PS agent in the mountain car problem as a function of the $\eta$ parameter is shown in Fig.~\ref{fig:PSetaMC}. As expected, for $\eta=0$, the performance is similar to the performance of the random agent, because all percept-action edges are internally enhanced starting from the time when they are visited. The best average performance of the PS agent is achieved for $\eta=0.024$, which is lower than the optimal value $\eta=0.2$ in the grid world problem. This is due to the fact that the shortest sequence of actions in the mountain car problem is roughly $10$ times longer and a stronger glow is required to remember the first actions in a longer sequence.

\begin{figure}[!ht]
	\centering
	\includegraphics[width=1\linewidth]{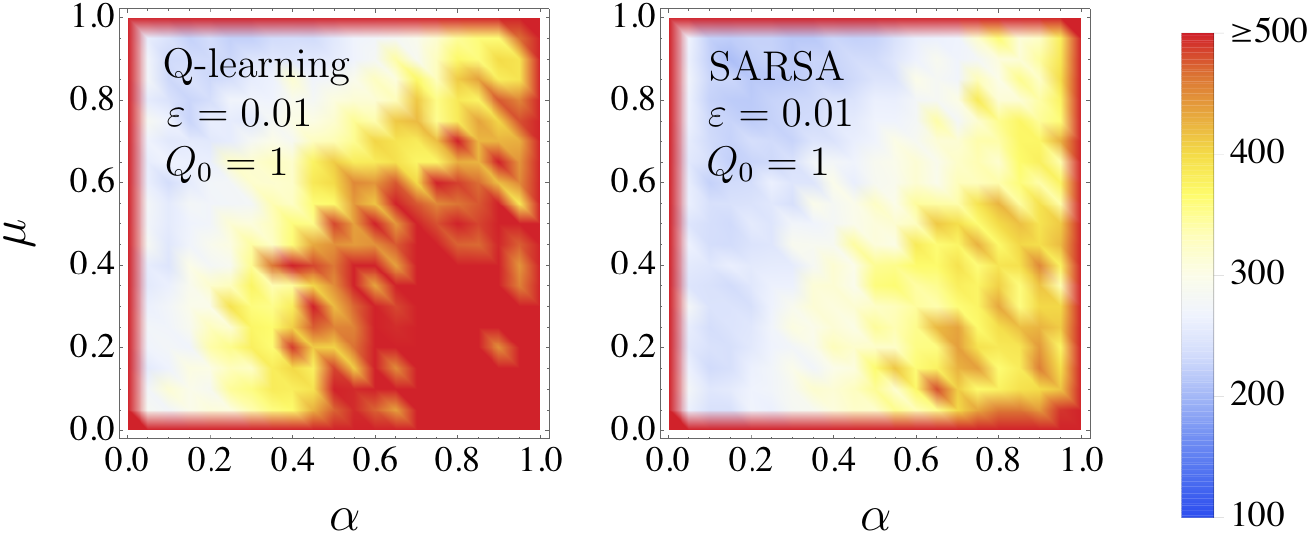}
	\caption{Q-learning (left) and SARSA (right) parameter optimization in the mountain car problem. The average number of steps (i.e., length of a trial) is calculated after $1000$ trials by averaging over $100$ agents. $21$ different values for the $\mu$ and $\alpha$ parameters are uniformly distributed in the $[0, 1]$ interval for fixed $Q_0$ and $\epsilon$. This leads to $4410$ different sets of parameters.}
	\label{fig:QSARSAparameter}
\end{figure}

The parameter optimization for the Q-learning and SARSA algorithms is more involved. As it requires significantly more computational time to optimize parameters in the mountain car problem than in the grid world problem, we sampled from the $\mu$ and $\alpha$ parameter intervals for values of $\epsilon = 0.01$ and $Q_0 = 1$. The dependence of the agents' performance on model parameters is shown in Fig.~\ref{fig:QSARSAparameter} for Q-learning (left) and SARSA (right). One can see that the region of optimal parameters is where the learning rate $\alpha$ is relatively small, which is different from the optimal range of the $\alpha$ parameter in the grid world problem (see Fig.~\ref{fig:QSARSA-GW}).

\begin{figure}[!ht]
	\centering
	\includegraphics[width=1\linewidth]{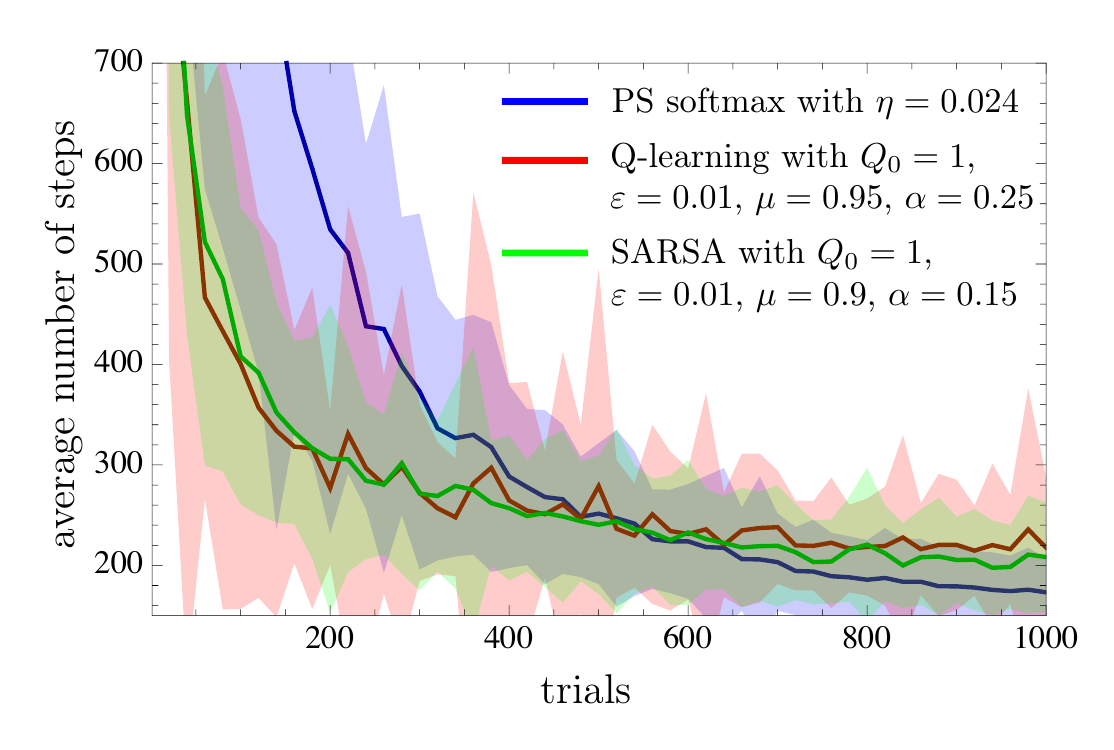}
	\caption{Learning curve of the PS, Q-learning and SARSA agents in the mountain car problem. The curves represent an average over $100$ agents of the same type. The shaded areas show mean squared deviation.}
	\label{fig:MClurves}
\end{figure}

Fig.~\ref{fig:MClurves} shows a comparison of learning curves of the PS (blue), Q-learning (red) and SARSA (green) agents, for the best set of model parameters that were found in each case. We see that the PS agent achieves the best result among all three types of agents with an average performance of $173.2$ steps and a mean squared deviation of $33.5$ steps. Q-learning and SARSA achieve a performance of $217.6$ and $208.3$ steps with mean squared deviation of $67.0$ and $54.2$ steps, respectively.

\section{Summary and Discussion} \label{sec:discussion}

In this paper, we studied the model of projective simulation, which we numerically tested in two benchmarking problems: the grid world and the mountain car problem. In both problems a reinforcement learning agent was tasked to navigate in the space of environment parameters -- $(x, y)$ in case of the grid world environment and $(x, v)$ in case of the mountain car environment. In both navigation problems we observed that the PS agent improves its performance after each trial, which is seen from simulations in Figs.~\ref{fig:GW-LearningCurves} and~\ref{fig:MClurves}. The optimal performance of the PS agent was obtained by choosing the glow parameter $\eta$, which is responsible for internally enhancing the weights of sequences of actions that lead to a reward. We have demonstrated that there is a straightforward way to choose this $\eta$ parameter, and gave heuristics of how the optimal $\eta$ depends on the maximum number of trials and the size of the problem. In addition, we showed that there is a tradeoff between the best average performance and the learning rate. As we demonstrated in Fig.~\ref{fig:EtaStudy}, both can be achieved by choosing a value for the $\eta$ parameter appropriately.

We have compared the performance of PS agent with Q-learning and SARSA in both benchmarking problems. The simulations of all $3$ types of agents suggest that the performance is qualitatively similar, as one can see in Figs.~\ref{fig:GW-LearningCurves} and~\ref{fig:MClurves}. Quantitatively, in the grid world problem we observed that Q-learning and SARSA converge to one of the optimal paths faster than the PS agent does. In the mountain car problem, we observed that the PS agent found a better solution compared to both Q-learning and SARSA. Importantly, the obtained performance of the Q-learning and SARSA agents was achieved by optimizing over $4$ parameters in the case of the grid world problem and $2$ parameters in the case of the mountain car problem. This can be compared to just $1$ parameter with a simple operational meaning in the case of PS. We demonstrated that optimizing parameters in the case of Q-learning and SARSA is much more demanding -- it required more than $54$ times more agent trials in the case of the grid world and $41$ times more agent trials in the case of the mountain car problem, which is a difference of more than one order of magnitude. Our results hence demonstrate that the PS agent is easy to set up, which is of importance in complex tasks where model parameter optimization is very costly.

\section*{Acknowledgment}
We wish to thank Vedran Dunjko for helpful discussions. This work was supported by the Austrian Science Fund (FWF) through Grant No. SFB FoQuS F4012, and by the Templeton World Charity Foundation (TWCF) through Grant No. TWCF0078/AB46.


\begin{thebibliography}{29}%
\makeatletter
\providecommand \@ifxundefined [1]{%
 \@ifx{#1\undefined}
}%
\providecommand \@ifnum [1]{%
 \ifnum #1\expandafter \@firstoftwo
 \else \expandafter \@secondoftwo
 \fi
}%
\providecommand \@ifx [1]{%
 \ifx #1\expandafter \@firstoftwo
 \else \expandafter \@secondoftwo
 \fi
}%
\providecommand \natexlab [1]{#1}%
\providecommand \enquote  [1]{``#1''}%
\providecommand \bibnamefont  [1]{#1}%
\providecommand \bibfnamefont [1]{#1}%
\providecommand \citenamefont [1]{#1}%
\providecommand \href@noop [0]{\@secondoftwo}%
\providecommand \href [0]{\begingroup \@sanitize@url \@href}%
\providecommand \@href[1]{\@@startlink{#1}\@@href}%
\providecommand \@@href[1]{\endgroup#1\@@endlink}%
\providecommand \@sanitize@url [0]{\catcode `\\12\catcode `\$12\catcode
  `\&12\catcode `\#12\catcode `\^12\catcode `\_12\catcode `\%12\relax}%
\providecommand \@@startlink[1]{}%
\providecommand \@@endlink[0]{}%
\providecommand \url  [0]{\begingroup\@sanitize@url \@url }%
\providecommand \@url [1]{\endgroup\@href {#1}{\urlprefix }}%
\providecommand \urlprefix  [0]{URL }%
\providecommand \Eprint [0]{\href }%
\providecommand \doibase [0]{http://dx.doi.org/}%
\providecommand \selectlanguage [0]{\@gobble}%
\providecommand \bibinfo  [0]{\@secondoftwo}%
\providecommand \bibfield  [0]{\@secondoftwo}%
\providecommand \translation [1]{[#1]}%
\providecommand \BibitemOpen [0]{}%
\providecommand \bibitemStop [0]{}%
\providecommand \bibitemNoStop [0]{.\EOS\space}%
\providecommand \EOS [0]{\spacefactor3000\relax}%
\providecommand \BibitemShut  [1]{\csname bibitem#1\endcsname}%
\let\auto@bib@innerbib\@empty
\bibitem [{\citenamefont {Russell}\ and\ \citenamefont
  {Norvig}(2010)}]{russell2010artificial}%
  \BibitemOpen
  \bibfield  {author} {\bibinfo {author} {\bibfnamefont {S.~J.}\ \bibnamefont
  {Russell}}\ and\ \bibinfo {author} {\bibfnamefont {P.}~\bibnamefont
  {Norvig}},\ }\href@noop {} {\emph {\bibinfo {title} {Artificial Intelligence:
  A Modern Approach}}},\ \bibinfo {edition} {3rd}\ ed.\ (\bibinfo  {publisher}
  {Prentice Hall, Englewood Cliffs},\ \bibinfo {year} {2010})\BibitemShut
  {NoStop}%
\bibitem [{\citenamefont {Briegel}\ and\ \citenamefont {De~las
  Cuevas}(2012)}]{briegel2012projective}%
  \BibitemOpen
  \bibfield  {author} {\bibinfo {author} {\bibfnamefont {H.~J.}\ \bibnamefont
  {Briegel}}\ and\ \bibinfo {author} {\bibfnamefont {G.}~\bibnamefont {De~las
  Cuevas}},\ }\href@noop {} {\bibfield  {journal} {\bibinfo  {journal} {Sci.
  Rep.}\ }\textbf {\bibinfo {volume} {2}},\ \bibinfo {pages} {400} (\bibinfo
  {year} {2012})}\BibitemShut {NoStop}%
\bibitem [{\citenamefont {Sutton}\ and\ \citenamefont
  {Barto}(1998)}]{sutton1998reinforcement}%
  \BibitemOpen
  \bibfield  {author} {\bibinfo {author} {\bibfnamefont {R.~S.}\ \bibnamefont
  {Sutton}}\ and\ \bibinfo {author} {\bibfnamefont {A.~G.}\ \bibnamefont
  {Barto}},\ }\href@noop {} {\emph {\bibinfo {title} {Reinforcement Learning:
  An Introduction}}}\ (\bibinfo  {publisher} {MIT press},\ \bibinfo {address}
  {Cambridge, MA, USA},\ \bibinfo {year} {1998})\BibitemShut {NoStop}%
\bibitem [{\citenamefont {Wiering}\ and\ \citenamefont {van
  Otterlo}(2012)}]{wiering2012reinforcement}%
  \BibitemOpen
  \bibfield  {author} {\bibinfo {author} {\bibfnamefont {M.}~\bibnamefont
  {Wiering}}\ and\ \bibinfo {author} {\bibfnamefont {M.~E.}\ \bibnamefont {van
  Otterlo}},\ }\href@noop {} {\emph {\bibinfo {title} {Reinforcement Learning:
  State-of-the-art}}},\ Vol.~\bibinfo {volume} {12}\ (\bibinfo  {publisher}
  {Springer},\ \bibinfo {year} {2012})\BibitemShut {NoStop}%
\bibitem [{\citenamefont {Mautner}\ \emph {et~al.}(2015)\citenamefont
  {Mautner}, \citenamefont {Makmal}, \citenamefont {Manzano}, \citenamefont
  {Tiersch},\ and\ \citenamefont {Briegel}}]{mautner2013projective}%
  \BibitemOpen
  \bibfield  {author} {\bibinfo {author} {\bibfnamefont {J.}~\bibnamefont
  {Mautner}}, \bibinfo {author} {\bibfnamefont {A.}~\bibnamefont {Makmal}},
  \bibinfo {author} {\bibfnamefont {D.}~\bibnamefont {Manzano}}, \bibinfo
  {author} {\bibfnamefont {M.}~\bibnamefont {Tiersch}}, \ and\ \bibinfo
  {author} {\bibfnamefont {H.~J.}\ \bibnamefont {Briegel}},\ }\href@noop {}
  {\bibfield  {journal} {\bibinfo  {journal} {New Gener. Comput.}\ }\textbf
  {\bibinfo {volume} {33}},\ \bibinfo {pages} {69} (\bibinfo {year}
  {2015})}\BibitemShut {NoStop}%
\bibitem [{\citenamefont {Hangl}\ \emph {et~al.}(2016)\citenamefont {Hangl},
  \citenamefont {Ugur}, \citenamefont {Szedmak},\ and\ \citenamefont
  {Piater}}]{simon2016}%
  \BibitemOpen
  \bibfield  {author} {\bibinfo {author} {\bibfnamefont {S.}~\bibnamefont
  {Hangl}}, \bibinfo {author} {\bibfnamefont {E.}~\bibnamefont {Ugur}},
  \bibinfo {author} {\bibfnamefont {S.}~\bibnamefont {Szedmak}}, \ and\
  \bibinfo {author} {\bibfnamefont {J.}~\bibnamefont {Piater}},\ }in\
  \href@noop {} {\emph {\bibinfo {booktitle} {Proc. IEEE/RSJ Int. Conf. Intell.
  Robots Syst.}}}\ (\bibinfo {year} {2016})\ pp.\ \bibinfo {pages}
  {2799--2804}\BibitemShut {NoStop}%
\bibitem [{\citenamefont {Hangl}\ \emph {et~al.}(2017)\citenamefont {Hangl},
  \citenamefont {Dunjko}, \citenamefont {Briegel},\ and\ \citenamefont
  {Piater}}]{simon2017skill}%
  \BibitemOpen
  \bibfield  {author} {\bibinfo {author} {\bibfnamefont {S.}~\bibnamefont
  {Hangl}}, \bibinfo {author} {\bibfnamefont {V.}~\bibnamefont {Dunjko}},
  \bibinfo {author} {\bibfnamefont {H.~J.}\ \bibnamefont {Briegel}}, \ and\
  \bibinfo {author} {\bibfnamefont {J.}~\bibnamefont {Piater}},\ }\href@noop {}
  {\bibfield  {journal} {\bibinfo  {journal} {arXiv:1706.08560}\ } (\bibinfo
  {year} {2017})}\BibitemShut {NoStop}%
\bibitem [{\citenamefont {Tiersch}\ \emph {et~al.}(2015)\citenamefont
  {Tiersch}, \citenamefont {Ganahl},\ and\ \citenamefont
  {Briegel}}]{tiersch2015adaptive}%
  \BibitemOpen
  \bibfield  {author} {\bibinfo {author} {\bibfnamefont {M.}~\bibnamefont
  {Tiersch}}, \bibinfo {author} {\bibfnamefont {E.~J.}\ \bibnamefont {Ganahl}},
  \ and\ \bibinfo {author} {\bibfnamefont {H.~J.}\ \bibnamefont {Briegel}},\
  }\href@noop {} {\bibfield  {journal} {\bibinfo  {journal} {Sci. Rep.}\
  }\textbf {\bibinfo {volume} {5}},\ \bibinfo {pages} {12874} (\bibinfo {year}
  {2015})}\BibitemShut {NoStop}%
\bibitem [{\citenamefont {Melnikov}\ \emph {et~al.}(2018)\citenamefont
  {Melnikov}, \citenamefont {Poulsen~Nautrup}, \citenamefont {Krenn},
  \citenamefont {Dunjko}, \citenamefont {Tiersch}, \citenamefont {Zeilinger},\
  and\ \citenamefont {Briegel}}]{melnikov2017active}%
  \BibitemOpen
  \bibfield  {author} {\bibinfo {author} {\bibfnamefont {A.~A.}\ \bibnamefont
  {Melnikov}}, \bibinfo {author} {\bibfnamefont {H.}~\bibnamefont
  {Poulsen~Nautrup}}, \bibinfo {author} {\bibfnamefont {M.}~\bibnamefont
  {Krenn}}, \bibinfo {author} {\bibfnamefont {V.}~\bibnamefont {Dunjko}},
  \bibinfo {author} {\bibfnamefont {M.}~\bibnamefont {Tiersch}}, \bibinfo
  {author} {\bibfnamefont {A.}~\bibnamefont {Zeilinger}}, \ and\ \bibinfo
  {author} {\bibfnamefont {H.~J.}\ \bibnamefont {Briegel}},\ }\href@noop {}
  {\bibfield  {journal} {\bibinfo  {journal} {Proc. Natl. Acad. Sci. U.S.A.}\
  }\textbf {\bibinfo {volume} {115}},\ \bibinfo {pages} {1221} (\bibinfo {year}
  {2018})}\BibitemShut {NoStop}%
\bibitem [{\citenamefont {Motwani}\ and\ \citenamefont
  {Raghavan}(1995)}]{randomized_algorithms_1995}%
  \BibitemOpen
  \bibfield  {author} {\bibinfo {author} {\bibfnamefont {R.}~\bibnamefont
  {Motwani}}\ and\ \bibinfo {author} {\bibfnamefont {P.}~\bibnamefont
  {Raghavan}},\ }\enquote {\bibinfo {title} {Randomized algorithms},}\ \
  (\bibinfo  {publisher} {Cambridge University Press},\ \bibinfo {address} {New
  York, USA},\ \bibinfo {year} {1995})\ Chap.~\bibinfo {chapter}
  {6}\BibitemShut {NoStop}%
\bibitem [{\citenamefont {Melnikov}\ \emph {et~al.}(2017)\citenamefont
  {Melnikov}, \citenamefont {Makmal}, \citenamefont {Dunjko},\ and\
  \citenamefont {Briegel}}]{melnikov2017projective}%
  \BibitemOpen
  \bibfield  {author} {\bibinfo {author} {\bibfnamefont {A.~A.}\ \bibnamefont
  {Melnikov}}, \bibinfo {author} {\bibfnamefont {A.}~\bibnamefont {Makmal}},
  \bibinfo {author} {\bibfnamefont {V.}~\bibnamefont {Dunjko}}, \ and\ \bibinfo
  {author} {\bibfnamefont {H.~J.}\ \bibnamefont {Briegel}},\ }\href@noop {}
  {\bibfield  {journal} {\bibinfo  {journal} {Sci. Rep.}\ }\textbf {\bibinfo
  {volume} {7}},\ \bibinfo {pages} {14430} (\bibinfo {year}
  {2017})}\BibitemShut {NoStop}%
\bibitem [{\citenamefont {Dunjko}\ \emph {et~al.}(2016)\citenamefont {Dunjko},
  \citenamefont {Taylor},\ and\ \citenamefont {Briegel}}]{dunjko2016quantum}%
  \BibitemOpen
  \bibfield  {author} {\bibinfo {author} {\bibfnamefont {V.}~\bibnamefont
  {Dunjko}}, \bibinfo {author} {\bibfnamefont {J.~M.}\ \bibnamefont {Taylor}},
  \ and\ \bibinfo {author} {\bibfnamefont {H.~J.}\ \bibnamefont {Briegel}},\
  }\href@noop {} {\bibfield  {journal} {\bibinfo  {journal} {Phys. Rev. Lett.}\
  }\textbf {\bibinfo {volume} {117}},\ \bibinfo {pages} {130501} (\bibinfo
  {year} {2016})}\BibitemShut {NoStop}%
\bibitem [{\citenamefont {Dunjko}\ \emph {et~al.}(2017)\citenamefont {Dunjko},
  \citenamefont {Taylor},\ and\ \citenamefont {Briegel}}]{dunjko2017advances}%
  \BibitemOpen
  \bibfield  {author} {\bibinfo {author} {\bibfnamefont {V.}~\bibnamefont
  {Dunjko}}, \bibinfo {author} {\bibfnamefont {J.~M.}\ \bibnamefont {Taylor}},
  \ and\ \bibinfo {author} {\bibfnamefont {H.~J.}\ \bibnamefont {Briegel}},\
  }in\ \href@noop {} {\emph {\bibinfo {booktitle} {Proc. IEEE Int. Conf. Syst.,
  Man, Cybern.}}}\ (\bibinfo {year} {2017})\ pp.\ \bibinfo {pages}
  {282--287}\BibitemShut {NoStop}%
\bibitem [{\citenamefont {Aharonov}\ \emph {et~al.}(1993)\citenamefont
  {Aharonov}, \citenamefont {Davidovich},\ and\ \citenamefont
  {Zagury}}]{aharonov1993}%
  \BibitemOpen
  \bibfield  {author} {\bibinfo {author} {\bibfnamefont {Y.}~\bibnamefont
  {Aharonov}}, \bibinfo {author} {\bibfnamefont {L.}~\bibnamefont
  {Davidovich}}, \ and\ \bibinfo {author} {\bibfnamefont {N.}~\bibnamefont
  {Zagury}},\ }\href@noop {} {\bibfield  {journal} {\bibinfo  {journal} {Phys.
  Rev. A}\ }\textbf {\bibinfo {volume} {48}},\ \bibinfo {pages} {1687}
  (\bibinfo {year} {1993})}\BibitemShut {NoStop}%
\bibitem [{\citenamefont {Aharonov}\ \emph {et~al.}(2001)\citenamefont
  {Aharonov}, \citenamefont {Ambainis}, \citenamefont {Kempe},\ and\
  \citenamefont {Vazirani}}]{aharonov2001quantum}%
  \BibitemOpen
  \bibfield  {author} {\bibinfo {author} {\bibfnamefont {D.}~\bibnamefont
  {Aharonov}}, \bibinfo {author} {\bibfnamefont {A.}~\bibnamefont {Ambainis}},
  \bibinfo {author} {\bibfnamefont {J.}~\bibnamefont {Kempe}}, \ and\ \bibinfo
  {author} {\bibfnamefont {U.}~\bibnamefont {Vazirani}},\ }in\ \href@noop {}
  {\emph {\bibinfo {booktitle} {Proceedings of the 33rd Annual ACM Symposium on
  Theory of Computing}}},\ \bibinfo {series and number} {STOC '01}\ (\bibinfo
  {year} {2001})\ pp.\ \bibinfo {pages} {50--59}\BibitemShut {NoStop}%
\bibitem [{\citenamefont {Venegas-Andraca}(2012)}]{venegas-andraca2012quantum}%
  \BibitemOpen
  \bibfield  {author} {\bibinfo {author} {\bibfnamefont {S.~E.}\ \bibnamefont
  {Venegas-Andraca}},\ }\href@noop {} {\bibfield  {journal} {\bibinfo
  {journal} {Quantum Information Processing}\ }\textbf {\bibinfo {volume}
  {11}},\ \bibinfo {pages} {1015} (\bibinfo {year} {2012})}\BibitemShut
  {NoStop}%
\bibitem [{\citenamefont {Paparo}\ \emph {et~al.}(2014)\citenamefont {Paparo},
  \citenamefont {Dunjko}, \citenamefont {Makmal}, \citenamefont
  {Martin-Delgado},\ and\ \citenamefont {Briegel}}]{paparo2014quantum}%
  \BibitemOpen
  \bibfield  {author} {\bibinfo {author} {\bibfnamefont {G.~D.}\ \bibnamefont
  {Paparo}}, \bibinfo {author} {\bibfnamefont {V.}~\bibnamefont {Dunjko}},
  \bibinfo {author} {\bibfnamefont {A.}~\bibnamefont {Makmal}}, \bibinfo
  {author} {\bibfnamefont {M.~A.}\ \bibnamefont {Martin-Delgado}}, \ and\
  \bibinfo {author} {\bibfnamefont {H.~J.}\ \bibnamefont {Briegel}},\
  }\href@noop {} {\bibfield  {journal} {\bibinfo  {journal} {Phys. Rev. X}\
  }\textbf {\bibinfo {volume} {4}},\ \bibinfo {pages} {031002} (\bibinfo {year}
  {2014})}\BibitemShut {NoStop}%
\bibitem [{\citenamefont {Dunjko}\ \emph {et~al.}(2015)\citenamefont {Dunjko},
  \citenamefont {Friis},\ and\ \citenamefont {Briegel}}]{dunjko2015quantum}%
  \BibitemOpen
  \bibfield  {author} {\bibinfo {author} {\bibfnamefont {V.}~\bibnamefont
  {Dunjko}}, \bibinfo {author} {\bibfnamefont {N.}~\bibnamefont {Friis}}, \
  and\ \bibinfo {author} {\bibfnamefont {H.~J.}\ \bibnamefont {Briegel}},\
  }\href@noop {} {\bibfield  {journal} {\bibinfo  {journal} {New J. Phys.}\
  }\textbf {\bibinfo {volume} {17}},\ \bibinfo {pages} {023006} (\bibinfo
  {year} {2015})}\BibitemShut {NoStop}%
\bibitem [{\citenamefont {Friis}\ \emph {et~al.}(2015)\citenamefont {Friis},
  \citenamefont {Melnikov}, \citenamefont {Kirchmair},\ and\ \citenamefont
  {Briegel}}]{friis2015coherent}%
  \BibitemOpen
  \bibfield  {author} {\bibinfo {author} {\bibfnamefont {N.}~\bibnamefont
  {Friis}}, \bibinfo {author} {\bibfnamefont {A.~A.}\ \bibnamefont {Melnikov}},
  \bibinfo {author} {\bibfnamefont {G.}~\bibnamefont {Kirchmair}}, \ and\
  \bibinfo {author} {\bibfnamefont {H.~J.}\ \bibnamefont {Briegel}},\
  }\href@noop {} {\bibfield  {journal} {\bibinfo  {journal} {Sci. Rep.}\
  }\textbf {\bibinfo {volume} {5}},\ \bibinfo {pages} {18036} (\bibinfo {year}
  {2015})}\BibitemShut {NoStop}%
\bibitem [{\citenamefont {Sriarunothai}\ \emph {et~al.}(2017)\citenamefont
  {Sriarunothai}, \citenamefont {W{\"o}lk}, \citenamefont {Giri}, \citenamefont
  {Friis}, \citenamefont {Dunjko}, \citenamefont {Briegel},\ and\ \citenamefont
  {Wunderlich}}]{sriarunothai2017speeding}%
  \BibitemOpen
  \bibfield  {author} {\bibinfo {author} {\bibfnamefont {T.}~\bibnamefont
  {Sriarunothai}}, \bibinfo {author} {\bibfnamefont {S.}~\bibnamefont
  {W{\"o}lk}}, \bibinfo {author} {\bibfnamefont {G.~S.}\ \bibnamefont {Giri}},
  \bibinfo {author} {\bibfnamefont {N.}~\bibnamefont {Friis}}, \bibinfo
  {author} {\bibfnamefont {V.}~\bibnamefont {Dunjko}}, \bibinfo {author}
  {\bibfnamefont {H.~J.}\ \bibnamefont {Briegel}}, \ and\ \bibinfo {author}
  {\bibfnamefont {C.}~\bibnamefont {Wunderlich}},\ }\href@noop {} {\bibfield
  {journal} {\bibinfo  {journal} {arXiv:1709.01366}\ } (\bibinfo {year}
  {2017})}\BibitemShut {NoStop}%
\bibitem [{\citenamefont {Rummery}\ and\ \citenamefont
  {Niranjan}(1994)}]{rummery1994line}%
  \BibitemOpen
  \bibfield  {author} {\bibinfo {author} {\bibfnamefont {G.~A.}\ \bibnamefont
  {Rummery}}\ and\ \bibinfo {author} {\bibfnamefont {M.}~\bibnamefont
  {Niranjan}},\ }\href@noop {} {\emph {\bibinfo {title} {On-Line {Q}-Learning
  Using Connectionist Systems}}},\ \bibinfo {type} {Tech. Rep.}\ \bibinfo
  {number} {CUED/F-INFENG/TR 166}\ (\bibinfo  {institution} {University of
  Cambridge},\ \bibinfo {year} {1994})\BibitemShut {NoStop}%
\bibitem [{\citenamefont {Sutton}(1990)}]{sutton1990integrated}%
  \BibitemOpen
  \bibfield  {author} {\bibinfo {author} {\bibfnamefont {R.~S.}\ \bibnamefont
  {Sutton}},\ }in\ \href@noop {} {\emph {\bibinfo {booktitle} {Proceedings of
  the 7th International Conference on Machine Learning}}}\ (\bibinfo {year}
  {1990})\ pp.\ \bibinfo {pages} {216--224}\BibitemShut {NoStop}%
\bibitem [{\citenamefont {Moore}(1990)}]{moore1990efficient}%
  \BibitemOpen
  \bibfield  {author} {\bibinfo {author} {\bibfnamefont {A.~W.}\ \bibnamefont
  {Moore}},\ }\href@noop {} {\emph {\bibinfo {title} {Efficient memory-based
  learning for robot control}}},\ \bibinfo {type} {Tech. Rep.}\ \bibinfo
  {number} {UCAM-CL-TR-209}\ (\bibinfo  {institution} {University of
  Cambridge},\ \bibinfo {year} {1990})\BibitemShut {NoStop}%
\bibitem [{\citenamefont {Melnikov}\ \emph {et~al.}(2014)\citenamefont
  {Melnikov}, \citenamefont {Makmal},\ and\ \citenamefont
  {Briegel}}]{melnikov2014projective}%
  \BibitemOpen
  \bibfield  {author} {\bibinfo {author} {\bibfnamefont {A.~A.}\ \bibnamefont
  {Melnikov}}, \bibinfo {author} {\bibfnamefont {A.}~\bibnamefont {Makmal}}, \
  and\ \bibinfo {author} {\bibfnamefont {H.~J.}\ \bibnamefont {Briegel}},\
  }\href@noop {} {\bibfield  {journal} {\bibinfo  {journal} {Artif. Intell.
  Res.}\ }\textbf {\bibinfo {volume} {3}},\ \bibinfo {pages} {24} (\bibinfo
  {year} {2014})},\ \bibinfo {note} {arXiv:1405.5459}\BibitemShut {NoStop}%
\bibitem [{\citenamefont {Makmal}\ \emph {et~al.}(2016)\citenamefont {Makmal},
  \citenamefont {Melnikov}, \citenamefont {Dunjko},\ and\ \citenamefont
  {Briegel}}]{makmal2016meta}%
  \BibitemOpen
  \bibfield  {author} {\bibinfo {author} {\bibfnamefont {A.}~\bibnamefont
  {Makmal}}, \bibinfo {author} {\bibfnamefont {A.~A.}\ \bibnamefont
  {Melnikov}}, \bibinfo {author} {\bibfnamefont {V.}~\bibnamefont {Dunjko}}, \
  and\ \bibinfo {author} {\bibfnamefont {H.~J.}\ \bibnamefont {Briegel}},\
  }\href@noop {} {\bibfield  {journal} {\bibinfo  {journal} {IEEE Access}\
  }\textbf {\bibinfo {volume} {4}},\ \bibinfo {pages} {2110} (\bibinfo {year}
  {2016})}\BibitemShut {NoStop}%
\bibitem [{\citenamefont {Ried}\ \emph {et~al.}(2017)\citenamefont {Ried},
  \citenamefont {M{\"u}ller},\ and\ \citenamefont
  {Briegel}}]{ried2017modelling}%
  \BibitemOpen
  \bibfield  {author} {\bibinfo {author} {\bibfnamefont {K.}~\bibnamefont
  {Ried}}, \bibinfo {author} {\bibfnamefont {T.}~\bibnamefont {M{\"u}ller}}, \
  and\ \bibinfo {author} {\bibfnamefont {H.~J.}\ \bibnamefont {Briegel}},\
  }\href@noop {} {\bibfield  {journal} {\bibinfo  {journal} {arXiv:1712.01334}\
  } (\bibinfo {year} {2017})}\BibitemShut {NoStop}%
\bibitem [{\citenamefont {Mirowski}\ \emph {et~al.}(2016)\citenamefont
  {Mirowski}, \citenamefont {Pascanu}, \citenamefont {Viola}, \citenamefont
  {Soyer}, \citenamefont {Ballard}, \citenamefont {Banino}, \citenamefont
  {Denil}, \citenamefont {Goroshin}, \citenamefont {Sifre}, \citenamefont
  {Kavukcuoglu}, \citenamefont {Kumaran},\ and\ \citenamefont
  {Hadsell}}]{mirowski2016learning}%
  \BibitemOpen
  \bibfield  {author} {\bibinfo {author} {\bibfnamefont {P.}~\bibnamefont
  {Mirowski}}, \bibinfo {author} {\bibfnamefont {R.}~\bibnamefont {Pascanu}},
  \bibinfo {author} {\bibfnamefont {F.}~\bibnamefont {Viola}}, \bibinfo
  {author} {\bibfnamefont {H.}~\bibnamefont {Soyer}}, \bibinfo {author}
  {\bibfnamefont {A.}~\bibnamefont {Ballard}}, \bibinfo {author} {\bibfnamefont
  {A.}~\bibnamefont {Banino}}, \bibinfo {author} {\bibfnamefont
  {M.}~\bibnamefont {Denil}}, \bibinfo {author} {\bibfnamefont
  {R.}~\bibnamefont {Goroshin}}, \bibinfo {author} {\bibfnamefont
  {L.}~\bibnamefont {Sifre}}, \bibinfo {author} {\bibfnamefont
  {K.}~\bibnamefont {Kavukcuoglu}}, \bibinfo {author} {\bibfnamefont
  {D.}~\bibnamefont {Kumaran}}, \ and\ \bibinfo {author} {\bibfnamefont
  {R.}~\bibnamefont {Hadsell}},\ }\href@noop {} {\bibfield  {journal} {\bibinfo
   {journal} {arXiv:1611.03673}\ } (\bibinfo {year} {2016})}\BibitemShut
  {NoStop}%
\bibitem [{\citenamefont {Mannucci}\ and\ \citenamefont {van
  Kampen}(2016)}]{hierarchical2016maze}%
  \BibitemOpen
  \bibfield  {author} {\bibinfo {author} {\bibfnamefont {T.}~\bibnamefont
  {Mannucci}}\ and\ \bibinfo {author} {\bibfnamefont {E.-J.}\ \bibnamefont {van
  Kampen}},\ }in\ \href@noop {} {\emph {\bibinfo {booktitle} {Proc. IEEE
  Symposium Series on Computational Intelligence}}}\ (\bibinfo {year}
  {2016})\BibitemShut {NoStop}%
\bibitem [{\citenamefont {Brockman}\ \emph {et~al.}(2016)\citenamefont
  {Brockman}, \citenamefont {Cheung}, \citenamefont {Pettersson}, \citenamefont
  {Schneider}, \citenamefont {Schulman}, \citenamefont {Tang},\ and\
  \citenamefont {Zaremba}}]{OpenAIpaper}%
  \BibitemOpen
  \bibfield  {author} {\bibinfo {author} {\bibfnamefont {G.}~\bibnamefont
  {Brockman}}, \bibinfo {author} {\bibfnamefont {V.}~\bibnamefont {Cheung}},
  \bibinfo {author} {\bibfnamefont {L.}~\bibnamefont {Pettersson}}, \bibinfo
  {author} {\bibfnamefont {J.}~\bibnamefont {Schneider}}, \bibinfo {author}
  {\bibfnamefont {J.}~\bibnamefont {Schulman}}, \bibinfo {author}
  {\bibfnamefont {J.}~\bibnamefont {Tang}}, \ and\ \bibinfo {author}
  {\bibfnamefont {W.}~\bibnamefont {Zaremba}},\ }\href@noop {} {\bibfield
  {journal} {\bibinfo  {journal} {arXiv:1606.01540}\ } (\bibinfo {year}
  {2016})}\BibitemShut {NoStop}%
\end{thebibliography}
\end{document}